\documentclass[10pt, a4paper]{article}

\usepackage[table]{xcolor}

\usepackage[]{lrec-coling2024} 

\usepackage{siunitx}
\usepackage{booktabs}
\usepackage{tikz}
\usetikzlibrary{tikzmark, calc}

\usepackage{graphicx}
\usepackage{subfig}
\usepackage{amssymb}
\usepackage{array}
\usepackage{inconsolata}
\usepackage{multirow}

\tikzset{
    double color fill/.code 2 args={
        \pgfdeclareverticalshading[%
            tikz@axis@top,tikz@axis@middle,tikz@axis@bottom%
        ]{diagonalfill}{100bp}{%
            color(0bp)=(tikz@axis@bottom);
            color(50bp)=(tikz@axis@bottom);
            color(50bp)=(tikz@axis@middle);
            color(50bp)=(tikz@axis@top);
            color(100bp)=(tikz@axis@top)
        }
        \tikzset{shade, left color=#1, right color=#2, shading=diagonalfill}
    }
}

\newcounter{BGnum}
\title{DrBenchmark: A Large Language Understanding Evaluation Benchmark for French Biomedical Domain}

\name{Yanis Labrak{\normalfont $^*$\textsuperscript{1,2}}, Adrien Bazoge{\normalfont $^*$\textsuperscript{3,4}}, Oumaima El Khettari{\normalfont\textsuperscript{4}} \\ {\large  \textbf{Mickael Rouvier{\normalfont\textsuperscript{1}}, Pacôme Constant dit Beaufils}{\normalfont\textsuperscript{5}}, \textbf{Natalia Grabar}{\normalfont\textsuperscript{6}}, \textbf{Béatrice Daille}{\normalfont\textsuperscript{4}} }\\ {\large\textbf{Solen Quiniou}{\normalfont\textsuperscript{4}}, \textbf{Emmanuel Morin}{\normalfont\textsuperscript{4}}, \textbf{Pierre-Antoine Gourraud}{\normalfont\textsuperscript{3}} and \textbf{Richard Dufour}{\normalfont\textsuperscript{1,4}}}}

\address{\textsuperscript{1}LIA, Avignon Université \hspace{4mm} \textsuperscript{2}Zenidoc \\ \textsuperscript{3}Nantes Université, CHU Nantes, Clinique des données, INSERM, CIC 1413, F-44000 Nantes, France \\
\textsuperscript{4}Nantes Université, École Centrale Nantes, CNRS, LS2N, UMR 6004, F-44000 Nantes, France \\
\textsuperscript{5}Nantes Université, CHU Nantes, Service de Neuroradiologie diagnostique et \\ interventionnelle, CNRS, INSERM, l'institut du thorax, F-44000 Nantes, France \\
\textsuperscript{6}UMR 8163 – STL CNRS, Université de Lille \\
         \{first.last\}@\{univ-avignon.fr, univ-nantes.fr, chu-nantes.fr, univ-lille.fr\} \\
}

\abstract{
The biomedical domain has sparked a significant interest in the field of Natural Language Processing (NLP), which has seen substantial advancements with pre-trained language models (PLMs). However, comparing these models has proven challenging due to variations in evaluation protocols across different models. A fair solution is to aggregate diverse downstream tasks into a benchmark, allowing for the assessment of intrinsic PLMs qualities from various perspectives. Although still limited to few languages, this initiative has been undertaken in the biomedical field, notably English and Chinese. This limitation hampers the evaluation of the latest French biomedical models, as they are either assessed on a minimal number of tasks with non-standardized protocols or evaluated using general downstream tasks. To bridge this research gap and account for the unique sensitivities of French, we present the first-ever publicly available French biomedical language understanding benchmark called DrBenchmark. It encompasses 20 diversified tasks, including named-entity recognition, part-of-speech tagging, question-answering, semantic textual similarity, and classification. We evaluate 8 state-of-the-art pre-trained masked language models (MLMs) on general and biomedical-specific data, as well as English specific MLMs to assess their cross-lingual capabilities. Our experiments reveal that no single model excels across all tasks, while generalist models are sometimes still competitive.
 \\ \newline  }

\begin{document}

\maketitleabstract

\section{Introduction}

For the past few years, the field of Natural Language Processing (NLP) has witnessed major breakthroughs, particularly in the area of language modeling. Newer approaches such as the self-attention mechanism~\cite{NIPS2017_3f5ee243}, Sparse Transformer~\cite{child2019generating}, and Replaced Token Detection~\cite{Clark2020ELECTRA} have emerged.
These advancements have enabled the application of language models pre-trained on large corpora of textual data to a wide range of NLP tasks.


The evaluation of proposed models and approaches is an essential step in verifying their quality and performance. In the context of pre-trained language models (PLMs), this validation typically involves assessing their performance on targeted downstream tasks. This task-selection process is crucial, as the performance of models can vary depending on the chosen ones. Consequently, a model that performs well in one context may deliver disappointing results in another one. To address this issue and validate the models' generalizability, evaluation benchmarks have emerged, typically encompassing diverse sets of tasks. Hence, the availability of evaluation benchmarks plays a vital role in driving continuous progress, fostering the development of community members, and facilitating fair comparison between models.

While numerous open benchmarks exist for general tasks in NLP across multiple languages, the biomedical field remains an area with relatively few proposed benchmarks, mainly for English and Chinese, facilitating the availability of many biomedical models in these two languages. Even if the gap in other languages is beginning to narrow with new specialized models, the development of evaluation platforms has been comparatively slower.



Although the French language is generally considered as well-endowed, it is notably lacking in evaluation resources within the biomedical field. 
To address this issue, we introduce DrBenchmark, the first comprehensive open benchmark for the French biomedical domain, comprising 20 diverse tasks.
These tasks encompass part-of-speech (POS) tagging, named-entity recognition (NER), classification, question-answering (QA), and semantic textual similarity (STS).

We also perform a quantitative study of 8 pre-trained state-of-the-art masked language models (MLMs) with different configurations (languages and domains) on DrBenchmark. Our in-depth analysis integrates a comparison of these models' performance, fine-tuning with limited data and word tokenization. Our main contributions are:




\begin{itemize}

  \item DrBenchmark, an original evaluation framework for French biomedical NLP domain aggregating a large set of 20 diversified, proven and challenging tasks.
  
  \item A quantitative study using our proposed biomedical benchmark on a wide range of 8 MLMs based on varied architectures, data sources and training strategies.
  
  \item The release under CC BY-SA 4.0 license on HuggingFace\footnote{\href{https://huggingface.co/DrBenchmark}{https://huggingface.co/DrBenchmark}} of a new open biomedical dataset with clinical cases manually annotated into the 22 International Classification of Diseases 10th Revision (ICD-10) categories.
  
  \item A modular, reproducible and easily customizable automated protocol using identical training and evaluation scripts allowing a simple and fair comparison, with, as input, only the evaluated language models. DrBenchmark is freely available under MIT license on GitHub, HuggingFace and summarized as a leaderboard on the website\footnote{\href{https://github.com/DrBenchmark/DrBenchmark}{https://github.com/DrBenchmark/DrBenchmark}}.
  
\end{itemize}

The paper is organized as follows: Section~\ref{sec:related-work} briefly introduces  historical NLP benchmarks, including biomedical ones. Section~\ref{sec:Overview} presents our proposed benchmark for French biomedical domain, with a focus on the downstream tasks. Section~\ref{sec:Experiments} presents the experimental protocol while Section~\ref{sec:benchmark-results} details the results and provides an analysis of the studied pre-trained models. Section~\ref{sec:conclusion} finally concludes the work and opens some perspectives.

\section{Related work}
\label{sec:related-work}

In the recent years, several NLP open benchmarks have been created to facilitate direct comparison between proposed approaches. Among the earliest benchmarks, DecaNLP~\cite{mccann2018natural} and GLUE~\cite{wang-etal-2018-glue}, focused on general English language understanding tasks rather than being specific to a particular domain. Thus, GLUE gathers nine tasks including text classification (linguistic acceptability, sentiment analysis, etc.), semantic analysis (paraphrase verification, sentence similarity, etc.), QA, coreference detection, and natural language inference (NLI). 
Following a similar concept, the French counterpart to GLUE, known as FLUE~\cite{le-etal-2020-flaubert-unsupervised}, consists of 7 general-domain tasks in French, covering areas such as text classification, paraphrasing, NLI, parsing, and word sense disambiguation. 

In the case of specialized domains, general benchmarks may not adequately evaluate the performance of in-domain models. Specifically, within the biomedical domain, only few benchmarks have been proposed, and they primarily focus on few languages. For instance, platforms like BLURB~\cite{Gu_2021} and BLUE~\cite{peng-etal-2019-transfer} predominantly offer benchmarks for English, while CBLUE~\cite{zhang-etal-2022-cblue} caters to the Chinese language. To provide more specific information, BLURB integrates 13 tasks, including NER, information and relation extraction, sentence similarity, text classification, and QA. BLUE encompasses 10 tasks, such as NER, sentence similarity, relation extraction, text classification, and inference. On the other hand, CBLUE covers 8 tasks, including NER, information extraction, text and intent classification, sentence similarity, and query relevance.

To our knowledge, aside the multilingual benchmark BigBIO~\citelanguageresource{fries2022bigbio} which includes only 4 corpora for French and is initially intended for generative text completion under zero-shot scenario, no large benchmark specialized in the French biomedical field exists. This makes the comparison of recent specialized models, such as \cite{labrak-etal-2023-drbert, touchent2023camembertbio, copara-etal-2020-contextualized, berhe-etal-2023-alibert}, extremely challenging.



\section{DrBenchmark Overview}
\label{sec:Overview}

\begin{table*}[t!]
\setlength\tabcolsep{12pt}

\resizebox{\textwidth}{!}{%
\setlength\extrarowheight{2.5pt}
\scriptsize
\centering
\begin{tabular}{ll|c|cccc}
\hline

\textbf{Dataset} & \textbf{Task} & \textbf{Metric} & \textbf{Train} & \textbf{Validation} & \textbf{Test} & \textbf{License} \\ \hline

CAS & POS tagging & SeqEval F1 & 2,653  & 379 & 758 & DUA \\ \hline
 
ESSAI & POS tagging & SeqEval F1 & 5,072   & 725 & 1,450 & DUA   \\ \hline
 
\multirow{2}{*}{QUAERO} & NER - EMEA & SeqEval F1 & 429 & 389 & 348 & GFDL 1.3   \\  
 & NER - MEDLINE & SeqEval F1 & 833   & 832 & 833 & GFDL 1.3  \\ \hline
 
\multirow{2}{*}{E3C} & NER - Clinical & SeqEval F1 & 969 & 140 & 293 & CC BY-NC  \\  
 & NER - Temporal & SeqEval F1 & 969 & 140 & 293 &  CC BY-NC \\ \hline
 
MorFITT & Multi-label Classification & Weighted F1 & 1514 & 1,022 & 1,088 & CC BY-SA 4.0  \\ \hline

\multirow{2}{*}{FrenchMedMCQA} & Question-Answering & Hamming / EMR & 2,171 & 312 & 622 & Apache 2.0 \\  
 & Multi-class Classification & Weighted F1 & 2,171 & 312 & 622 & Apache 2.0  \\ \hline 
 
\multirow{3}{*}{Mantra-GSC} & NER - EMEA & SeqEval F1 & 70  & 10 & 20 &  CC BY 4.0 \\  
 & NER - Medline & SeqEval F1 & 70  & 10 & 20 &  CC BY 4.0 \\  
 & NER - Patents & SeqEval F1 & 35 & 5 & 10 &  CC BY 4.0 \\ \hline 
 
CLISTER & Semantic Textual Similarity & EDRM / Spearman & 499 & 101 & 400 &  DUA \\ \hline


\multirow{2}{*}{DEFT-2020} & Semantic Textual Similarity & EDRM / Spearman & 498 & 102 & 410 &  DUA \\  
& Multi-class Classification & Weighted F1 & 460 & 112 & 530 & DUA  \\ 
\hline

\multirow{2}{*}{DEFT-2021} & Multi-label Classification & Weighted F1 & 118 & 49 & 108 &  DUA \\
& NER & SeqEval F1 & 2,153   & 793 & 1,766 &  DUA \\ \hline 

\multirow{1}{*}{DiaMed} & Multi-class Classification & Weighted F1 & 509 & 76 & 154 & CC BY-SA 4.0  \\ \hline

\multirow{2}{*}{PxCorpus} & NER & SeqEval F1 & 1,386 & 198 & 397 &  CC BY 4.0 \\  
 & Multi-class Classification & Weighted F1 & 1,386   & 198 & 397 &  CC BY 4.0 \\ \hline 
 
\end{tabular}
}
\caption{Descriptions and statistics of the 20 tasks included in DrBenchmark.}
\label{table:description}
\end{table*}

Our proposed benchmark comprises 20 French biomedical language understanding tasks, one of which is specifically created for this benchmark. The descriptions and statistics of these tasks are presented in Table~\ref{table:description}. DrBenchmark encompasses the following overall aspects:

\begin{enumerate}
  \item \textbf{A variety of tasks with different requirements and objectives:} Part-of-Speech (POS) tagging, Multi-class, Multi-label and Intent classification, Named-Entity Recognition (NER), Multiple-Choice Question-Answering (MCQA), and Semantic Textual Similarity (STS).
  
  \item \textbf{A diverse range of data origins:} Scientific literature, clinical trials, clinical cases, speech transcriptions, and more as described in Table~\ref{tab:data_sources}.
  
\end{enumerate}

Please note that within DrBenchmark, we include classical tasks like NER and POS tagging, as well as more specific and challenging tasks like MCQA and multi-label classification. In Section~\ref{sec:Tasks}, we provide an overview of the different French downstream tasks, while, in Section~\ref{sec:Reproducibility}, we offer insights into the pipeline and its reproducibility.

\begin{table}[!htb]
\resizebox{\columnwidth}{!}{%
\setlength\extrarowheight{1pt}
\centering
\begin{tabular}{cc}
\hline
\textbf{Dataset}       & \textbf{Sources}                   \\ \hline
\textit{CAS}           & Clinical cases                     \\
\textit{ESSAI}         & Clinical trial protocols           \\
\textit{QUAERO}        & Drug leaflets \& Biomedical titles \\
\textit{E3C}           & Clinical cases                     \\
\textit{MorFITT}       & Biomedical abstracts               \\
\textit{FrenchMedMCQA} & Pharmacy Exam                      \\
\textit{Mantra-GSC}    & Biomedical abstract / titles, drug labels, \& patent \\
\textit{CLISTER}       & Clinical cases                     \\
\textit{DEFT-2020}     & Clinical cases, encyclopedia \& drug labels\\
\textit{DEFT-2021}     & Clinical cases                     \\
\textit{DiaMed}        & Clinical cases                     \\
\textit{PxCorpus}      & Drug prescriptions transcripts     \\ \hline
\end{tabular}%
}
\caption{Data sources covered by each datasets.}
\label{tab:data_sources}
\end{table}

\subsection{Downstream tasks}
\label{sec:Tasks}

\paragraph{DEFT-2020}~\citelanguageresource{cardon-etal-2020-presentation} contains clinical cases, encyclopedia and drug labels introduced in the 2020 edition of an annual French Text Mining Challenge, called DEFT, and annotated for two tasks: (i) textual similarity and (ii) multi-class classification. The first task aims at identifying the degree of similarity within pairs of sentences, from 0 (the less similar) to 5 (the most similar). The second task consists in identifying, for a given sentence, the most similar sentence among three sentences provided.

\paragraph{DEFT-2021}~\citelanguageresource{grouin-etal-2021-classification} is a subset of 275 clinical cases taken from the 2019 edition of DEFT. This dataset is manually annotated in two tasks: (i) multi-label classification and (ii) NER. The multi-label classification task focuses on identifying the patient's clinical profile based on the diseases, signs, or symptoms mentioned in the clinical cases. The dataset is annotated with 23 axes from Chapter C of the Medical Subject Headings (MeSH). The second task involves fine-grained information extraction for 13 types of entities (more detail in Appendix~\ref{sec:classes-DEFT2021}).


\paragraph{E3C}~\citelanguageresource{Magnini2020TheEP} is a multilingual dataset of clinical cases annotated for the NER task. It consists of two types of annotations (more detail in Appendix~\ref{sec:classes-E3C}): (i) clinical entities (e.g., pathologies), (ii) temporal information and factuality (e.g., events). While the dataset covers 5 languages, only the French portion is retained for the benchmark. Since the dataset does not come with pre-defined subsets, we performed a 70 / 10 / 20 random split, as described in Table~\ref{table:sources-e3c}.


\begin{table}[!htb]
\setlength\tabcolsep{3.5pt}
\setlength\extrarowheight{1pt}
\scriptsize
\centering
\begin{tabular}{c|ccc}
\hline
\textbf{Subset}   & \textbf{Train}      & \textbf{Validation}        & \textbf{Test} \\ \hline
\textbf{Clinical} & 87.38 \% of layer 2 & 12.62 \% of layer 2 & 100 \% of layer 1 \\ \hline
\textbf{Temporal} & 70 \% of layer 1    & 10 \% of layer 1    & 20 \% of layer 1 \\ \hline
\end{tabular}%
\caption{Description of the sources for E3C.}
\label{table:sources-e3c}
\end{table}


\paragraph{The QUAERO French Medical Corpus}~\citelanguageresource{Nvol2014TheQF}, simply referred to as QUAERO in this paper, contains annotated entities and concepts for NER tasks. The dataset covers two text genres (drug leaflets and biomedical titles), consisting of a total of 103,056 words sourced from EMEA or MEDLINE. 10 entity categories corresponding to the UMLS Semantic Groups~\citelanguageresource{Lindberg-MIM1993} were annotated (more detail in Appendix~\ref{sec:classes-QUAERO})..
In total, 26,409 entity annotations were mapped to 5,797 unique UMLS concepts. 
Due to the presence of nested entities in annotations, we simplified the evaluation process by retaining only annotations at the higher granularity level from the BigBio~\citelanguageresource{fries2022bigbio} implementation, following the approach described in~\citet{touchent2023camembertbio}, which translates into an average loss of 6.06\% of the annotations on EMEA and 8.90\% on MEDLINE. Additionally, considering that some documents from EMEA exceed the maximum input sequence length that most current language models can handle, we decided to split these documents into sentences.

\paragraph{MorFITT}~\citelanguageresource{labrak:hal-04125879} is a multi-label dataset annotated with medical specialties. It contains 3,624 biomedical abstracts from PMC Open Access. It has been annotated across 12 medical specialties (more detail in Appendix~\ref{sec:classes-MorFITT}), for a total of 5,116 annotations.

\paragraph{FrenchMedMCQA}~\citelanguageresource{labrak:hal-03824241} is a Multiple-Choice Question-Answering (MCQA) dataset for biomedical domain. It contains 3,105 questions coming from real exams of the French medical specialization diploma in pharmacy, integrating single and multiple answers. The first task consists of automatically identifying the set of correct answers among the 5 proposed for a given question. The second task consists of identifying the number of answers (between 1 and 5) supposedly correct for a given question.

\paragraph{Mantra-GSC}~\citelanguageresource{10.1093/jamia/ocv037} is a multilingual dataset annotated for biomedical NER. From the 5 languages covered, we included only the French subset in this benchmark. The dataset is obtained from 3 sources which have been partitioned to be evaluated separately by 2 annotation schemes (more detail in Appendix \ref{sec:classes-MantraGSC}): Medline (11 classes), and EMEA and Patents (10 classes). The sources cover different types of documents (biomedical abstracts/titles, drug labels and patents). To ensure evaluation consistency, we randomly split the dataset into 3 subsets: 70\% for training, 10\% for validation, and 20\% for testing.

\paragraph{CLISTER}~\citelanguageresource{hiebel-etal-2022-clister-corpus} is a French clinical cases Semantic textual similarity (STS) dataset of 1,000 sentence pairs manually annotated by several annotators, who assigned similarity scores ranging from 0 to 5 to each pair. The scores were then averaged together to obtain a floating-point number representing the overall similarity. The objective of this dataset is to develop models that can automatically predict a similarity score that closely aligns with the reference score based solely on the two sentences provided. 


\paragraph{CAS}~\citelanguageresource{grabar:hal-01937096} comprises 3,790 clinical cases that have been annotated for POS tagging with 31 classes using automatic annotations through Tagex~\footnote{\href{https://allgo.inria.fr/app/tagex}{https://allgo.inria.fr/app/tagex}}, with an evaluation conducted by comparing the automatic outputs against manual annotations. This evaluation yielded 98\% precision.  Since the dataset does not come with predefined subsets, we made the decision to randomly split it into 3 subsets of 70\%, 10\% and 20\% of the total data for training, validation and test respectively.

\paragraph{ESSAI}~\citelanguageresource{dalloux_claveau_grabar_oliveira_moro_gumiel_carvalho_2021} contains 7,247 clinical trial protocols annotated in 41 POS tags using TreeTagger \cite{Schmid1994}. As the dataset was not originally divided into 3 subsets, we applied the same procedure as on the CAS corpus.

\paragraph{PxCorpus}~\citelanguageresource{Kocabiyikoglu2022} is a spoken language understanding dataset in the domain of medical drug prescription transcripts. It includes 4 hours (1,981 recordings) of transcribed and annotated dialogues focused on drug prescriptions. The recordings were manually transcribed and semantically annotated. The first task involves classifying the textual utterances into one of the 4 intent classes (prescribe, replace, negate, none). The second task is a NER task where each word in a sequence is classified into one of 38 classes, such as drug, dose, or mode (more detail in Appendix \ref{sec:classes-pxcorpus}).

\paragraph{DiaMed} is an original dataset created specifically for DrBenchmark. It comprises 739 new French clinical cases collected from an open source journal (The Pan African Medical Journal). The cases have been manually annotated by several annotators, one of which is a medical expert, into 22 chapters of the International Classification of Diseases, 10th Revision (ICD-10)~\citelanguageresource{WorldHealthOrganization2019}. These chapters provide a general description of the type of injury or disease. To ease the annotation process, only label at the chapter level were used (more detail in Appendix \ref{sec:classes-DiaMed}). The inter-annotator agreement between the 4 annotators has been computed for two annotation sessions (see Table~\ref{tab:my-table}), with 15 different clinical cases assessed per session. 

\begin{table}[htb!]
\scriptsize
\centering
\setlength\tabcolsep{3pt}
\setlength\extrarowheight{1.7pt}
\begin{tabular}{l|rr|rr}
\hline
& \multicolumn{2}{c|}{Session 1 - 0 to 15 docs}                           & \multicolumn{2}{c}{Session 2 - 15 to 30 docs}                           \\ \hline
Annotator ID & \multicolumn{1}{c|}{$\kappa$} & \multicolumn{1}{c|}{$\mathcal{G}$} & \multicolumn{1}{c|}{$\kappa$} & \multicolumn{1}{c}{$\mathcal{G}$} \\ \hline
Annotator 1 \& 2 & \multicolumn{1}{r|}{0.538} & 0.566 & \multicolumn{1}{r|}{0.697} & 0.705 \\ 
Annotator 1 \& 3 & \multicolumn{1}{r|}{0.682}       & 0.709                         & \multicolumn{1}{r|}{0.697}       & 0.705                         \\ 
Annotator 1 \& 4 & \multicolumn{1}{r|}{0.397}       & 0.429                         & \multicolumn{1}{r|}{0.548}       & 0.558                         \\ 
Annotator 2 \& 3 & \multicolumn{1}{r|}{0.311}       & 0.357                         & \multicolumn{1}{r|}{1.000}         & 1.000                           \\ 
Annotator 2 \& 4 & \multicolumn{1}{r|}{0.472}       & 0.497                         & \multicolumn{1}{r|}{0.672}       & 0.707                         \\ 
Annotator 3 \& 4 & \multicolumn{1}{r|}{0.311}       & 0.354                         & \multicolumn{1}{r|}{0.672}       & 0.707                         \\ 
\hline
Average & \multicolumn{1}{r|}{0.452}       & 0.485                         & \multicolumn{1}{r|}{0.714} & 0.730 \\ \hline
\end{tabular}%
\caption{Inter-annotator agreement statistics. $\kappa$ is referring to Kappa Cohen and $\mathcal{G}$ to Gwet's AC1.}
\label{tab:my-table}
\end{table}








\begin{table*}[!htb]
\setlength\tabcolsep{2.5pt}
\setlength\extrarowheight{3pt}
\scriptsize
\centering
\begin{tabular}{llccccc}
\hline
\textbf{} & \textbf{Model} & \textbf{Tokenizer} & \textbf{Vocabulary} & \textbf{Pretraining} & \textbf{Corpus} & \textbf{Text Size} \\ \hline

\multirow{3}{*}{French Generalist}
& CamemBERTa & SentencePiece 32K & CCNET & from-scratch & CCNET & 4 GB \\
& CamemBERT & SentencePiece 32K & OSCAR & from-scratch & OSCAR & 138 GB \\
& FlauBERT & BPE 50K & Wiki + Web crawl & from-scratch & Wiki + Web crawl & 71 GB \\ \hline

 \multirow{3}{*}{French Biomedical}
& DrBERT-FS & SentencePiece 32K & NACHOS & from-scratch & NACHOS & 7.4 GB \\
& DrBERT-CP & WordPiece 30K & PubMed & continual pretraining & PubMed + NACHOS   & 21 + 4 GB           \\
& CamemBERT-bio & SentencePiece 32K & OSCAR & continual pretraining & OSCAR + biomed-fr & 138 + 2.7 GB \\ \hline

\multirow{1}{*}{Cross-lingual Generalist}
& XLM-RoBERTa & WordPiece 30K & CC-100 & from-scratch & CC-100 & 2.5 TB \\ \hline

\multirow{1}{*}{English Biomedical} & PubMedBERT & WordPiece 30K & PubMed & from-scratch & PubMed & 21 GB \\
 \hline
 
\end{tabular}%
\caption{Summary of the pre-training specifications for the different BERT-based models compared.
}
\label{table:baselines}
\end{table*}

\subsection{Reproducibility and usage}
\label{sec:Reproducibility}

To facilitate the adoption of DrBenchmark and ensure consistency in implementations, we have developed a practical toolkit based on the HuggingFace Datasets library~\cite{lhoest-etal-2021-datasets}. This toolkit includes data loaders that adhere to normalized schemes and predefined data splits. It also provides pre-training and evaluation scripts for each of the tasks, utilizing the HuggingFace Transformers~\cite{wolf-etal-2020-transformers} and PyTorch~\cite{10.5555/3454287.3455008} libraries. For further guidance, we have integrated all the training details, including hyperparameters, in Appendix~\ref{sec:hyperparam}. This information will help users to reproduce and customize the experiments conducted with DrBenchmark.

\section{Experimental Protocol}
\label{sec:Experiments}


In this section, we outline the experimental protocol used to compare the performance of existing language models within DrBenchmark. To guarantee fair comparison, we focus exclusively on pre-trained masked language models (MLMs) in this study. These MLMs are based on BERT-like architectures~\cite{devlin-etal-2019-bert}.

We first provide a brief overview in Section~\ref{s:mlms} of the 8 pre-trained language models that were studied: French generalist models (CamemBERT, CamemBERTa and FlauBERT), cross-lingual generalist model (XLM-RoBERTa), French biomedical models (DrBERT and CamemBERT-bio), and English biomedical model (PubMedBERT). Subsequently, in Section~\ref{sec:evaluation}, we describe the evaluation protocol employed to assess the performance of these models.

\subsection{Pre-trained Masked Language Models}
\label{s:mlms}

Table~\ref{table:baselines} summarizes the models and their parameters compared on DrBenchmark.

\paragraph{CamemBERT}~\cite{martin-etal-2020-camembert} is a RoBERTa based model for French, pre-trained from-scratch on the generalist French 138 GB subset of \texttt{OSCAR} corpus~\citelanguageresource{OrtizSuarezSagotRomary2019}.

\paragraph{CamemBERTa}~\cite{antoun:hal-03963729} is a DeBERTaV3~\cite{he2023debertav} based model pre-trained from-scratch on around 30\% of the French subset of \texttt{CCNET} corpus~\citelanguageresource{wenzek-etal-2020-ccnet} used for CamemBERT$_{CCNET}$, that had seen approximately 133 billion tokens during its pre-training.

\paragraph{FlauBERT}~\cite{le-etal-2020-flaubert-unsupervised} is a BERT based model pre-trained from-scratch using a subsample of 71 GB of the French Common Crawl and Wikipedia corpora.

\paragraph{XLM-RoBERTa}~\cite{conneau-etal-2020-unsupervised} is a cross-lingual RoBERTa based model trained on 116 languages, including French, by using 2.5 TB of the CommonCrawl corpus.



\paragraph{PubMedBERT}~\cite{Gu_2021} is a BERT based biomedical-specific model pre-trained from-scratch on the 3.1 billion words of the \texttt{PubMed} corpus (21 GB). This is the only model for English.

\paragraph{DrBERT-FS and DrBERT-CP}~\cite{labrak-etal-2023-drbert} are French biomedical MLMs built using a from-scratch pre-training of RoBERTa (DrBERT-FS) and continual pre-training of PubMedBERT (DrBERT-CP) from the French public biomedical corpus \texttt{NACHOS}~\cite{labrak-etal-2023-drbert} integrating 1.08 billion words (7.4 GB) and 646 million words (4 GB) respectively. 

\paragraph{CamemBERT-bio}~\cite{touchent2023camembertbio} is a French biomedical language model built using a continual pre-training of the CamemBERT$_{OSCAR-138GB}$ model. It was trained on the French public corpus \texttt{biomed-fr}~\cite{touchent2023camembertbio} with 413 million words (2.7 GB) and a wide range of data collected on the web.


\subsection{Models evaluation}
\label{sec:evaluation}

\begin{table*}[t!]
\setlength\tabcolsep{2.0pt}
\setlength\extrarowheight{4.0pt}
\scriptsize
\centering
\resizebox{\textwidth}{!}{%
\begin{tabular}{ll|c|ccc|ccc|c|c}
\hline
\textbf{} &
  \textbf{} &
  &
  \multicolumn{3}{c|}{\textbf{French Generalist}} &
  \multicolumn{3}{c|}{\textbf{French Biomedical}} &
  \multicolumn{1}{c|}{\textbf{English Biomedical}} &
  \textbf{Cross-lingual Generalist} \\ \hline
\textbf{Dataset} &
  \textbf{Task} &
  \textbf{Baseline} &
  \textbf{CamemBERT} &
  \textbf{CamemBERTa} &
  \textbf{FlauBERT} &
  \textbf{DrBERT-FS} &
  \textbf{DrBERT-CP} &
  \textbf{CamemBERT-bio} &
  \textbf{PubMedBERT} &
  \textbf{XLM-RoBERTa} \\ \hline
CAS & POS & 23.50 & 95.53** & 96.56** & 95.22** & \textbf{96.93} & 96.46** & 95.22** & 94.82** & \underline{96.91} \\
  \hline
ESSAI & POS & 26.31 & 97.38** & 98.08** & 97.05* & \textbf{98.41} & 98.01** & 97.39** & 97.42** & \underline{98.34} \\ 
 \hline
\multirow{2}{*}{QUAERO} &
  NER EMEA &
  8.37 &
  62.68** &
  64.86** &
  \textbf{74.86} &
  64.11** &
  \underline{67.05}** &
  66.59** &
  53.19** &
  64.47** \\
 &
  NER MEDLINE &
  4.92 &
  55.25** &
  55.60** &
  48.98 &
  55.82** &
  \textbf{60.10} &
  \underline{58.94} &
  53.26** &
  51.12** \\ \hline
\multirow{2}{*}{E3C} &
  NER Clinical &
  4.47 &
  54.70** &
  55.53 &
  47.61 &
  54.45 &
  \underline{56.55} &
  \textbf{56.96} &
  38.34 &
  52.87** \\
 &
  NER Temporal &
  21.74 &
  \textbf{83.45} &
  83.22 &
  61.64 &
  81.48** &
  83.43 &
  \underline{83.44} &
  80.86** &
  82.6 \\ \hline
MorFITT &
  Multi-Label CLS &
  3.24 &
  64.21** &
  66.28** &
  \underline{70.25} &
  68.70** &
  \textbf{70.99} &
  67.53** &
  68.58** &
  67.28** \\ \hline
\multirow{2}{*}{FrenchMedMCQA} &
  MCQA &
  21.83 / \textbf{11.57} &
  28.53 / 2.25** &
  29.77 / 2.57** &
  27.88 / 2.09** &
  31.07 / \underline{3.22}** &
  32.41 / 2.89** &
  \textbf{35.3} / 1.45 &
  32.90 / 1.61** &
  \underline{34.74} / 2.09** \\
 &
  CLS &
  8.37 &
  \underline{66.21} &
  64.44** &
  61.88 &
  65.38 &
  \textbf{66.22} &
  65.79 &
  65.41* &
  64.69* \\ \hline
\multirow{3}{*}{MantraGSC} &
  NER FR EMEA &
  0.00 &
  29.14** &
  40.84** &
  \underline{66.20} &
  \textbf{66.23} &
  60.88 &
  30.63** &
  40.14** &
  52.64* \\
 &
  NER FR Medline &
  7.78 &
  23.20** &
  22.55** &
  20.69 &
  \textbf{42.38} &
  \underline{35.52} &
  23.66** &
  27.53* &
  18.73* \\
 &
  NER FR Patents &
  6.20 &
  00.00** &
  \underline{44.16}** &
  31.47** &
  \textbf{57.34} &
  39.68 &
  00.00** &
  4.51** &
  8.58** \\ \hline
CLISTER &
  STS &
  0.44 / 0.00 &
  0.55 / 0.33** &
  0.56 / 0.47** &
  0.50 / 0.29** &
  \underline{0.62} / \underline{0.57}** &
  0.60 / 0.49* &
  0.54 / 0.26** &
  \textbf{0.70} / \textbf{0.78} &
  0.49 / 0.23** \\ \hline
\multirow{2}{*}{DEFT-2020} &
  STS &
  0.49 / 0.00 &
  0.59 / 0.58** &
  0.59 / 0.43** &
  0.58 / 0.51** &
  0.72 / \underline{0.81}* &
  \underline{0.73} / \textbf{0.86} &
  0.58 / 0.32** &
  \textbf{0.78} / \textbf{0.86} &
  0.60 / 0.26** \\
 &
  CLS &
  14.00 &
  \underline{96.31} &
  \textbf{97.96} &
  42.37** &
  82.38 &
  95.71* &
  94.78* &
  95.33* &
  67.66** \\ \hline
\multirow{2}{*}{DEFT-2021} &
  Multi-Label CLS &
  24.49 &
  18.04** &
  18.04** &
  \textbf{39.21} &
  \underline{34.15}** &
  30.04** &
  17.82** &
  25.53** &
  24.46** \\
 &
  NER &
  0.00 &
  62.76** &
  62.61** &
  33.51 &
  60.44** &
  \underline{63.43}* &
  \textbf{64.36} &
  60.27** &
  60.32** \\ \hline
DiaMED &
  CLS &
  15.36 &
  30.40** &
  24.05** &
  34.08** &
  \textbf{60.45} &
  54.43** &
  39.57** &
  \underline{54.96}** &
  26.69** \\ \hline
\multirow{2}{*}{PxCorpus} &
  NER &
  10.00 &
  92.89** &
  95.05** &
  47.57 &
  \textbf{95.88} &
  71.38 &
  93.08** &
  94.66** &
  \underline{95.80} \\
 &
  CLS &
  84.78 &
  94.41 &
  93.95 &
  93.45* &
  94.43 &
  \textbf{94.52} &
  \underline{94.49} &
  93.12 &
  93.91 \\ \hline 
\end{tabular}%
}
\caption{Performance of the studied models over 4 runs. Best model in bold and second is underlined. Statistical
significance is computed using Student’s t-test: * stands for p < 0.05, ** stands for p < 0.01.}
\label{table:results}
\end{table*}

All the models are fine-tuned regarding a strict protocol using the same hyperparameters for each downstream task. The reported results are obtained by averaging the scores from four separate runs, thus ensuring robustness and reliability. We also report statistical significance computed using Student’s t-test.

To ensure a fair and consistent comparison among systems for sequence-to-sequence tasks such as POS tagging and NER, we chose the SeqEval~\cite{seqeval} metric in conjunction with the IOB2 format and the training of all the models to predict only the label on the first token of each word as mentioned by~\citet{touchent2023camembertbio}. It provides a tokenizer-agnostic evaluation and mitigates any correlation between models' performances and the tokenization process.

For STS tasks, the models' performance was assessed using two metrics: (1) the Spearman correlation, and (2) the mean relative solution distance accuracy (EDRM), as defined by the original authors of the DEFT-2020 dataset~\citelanguageresource{cardon-etal-2020-presentation}.



\section{Experiments and Results}
\label{sec:benchmark-results}

In Section~\ref{mlm-performances}, we compare the results obtained by each model within DrBenchmark, which permits to position a wide range of state-of-the-art models in the biomedical field across various NLP tasks. Then, we propose to gain a comprehensive understanding of the models' behavior by examining areas such as low-resource fine-tuning scenarios (Section~\ref{s:fine-tune}) and the analysis of word tokenization of the studied models (Section~\ref{s:word_tok}).


\subsection{Comparison of models performance}
\label{mlm-performances}

The results of the 8 models are reported in Table~\ref{table:results} and compared to a baseline obtained by considering the majority class for all predictions. Overall, although we might anticipate certain models to excel in all tasks, we discovered that no single model outperforms the rest in all application scenarios. Interestingly, most of the models examined manage to secure the top position in at least one of the French biomedical downstream tasks studied. The only exception pertains to the cross-lingual generalist model (\texttt{XLM-RoBERTa}), which manages to reach the second-best position on several tasks.

Despite this unexpected outcome, we observe that French biomedical language models (\texttt{DrBERT-FS}, \texttt{DrBERT-CP}, \texttt{CamemBERT-bio}), presumed to be the most aligned with the nature of the data of the benchmark, exhibit indeed superior performance across many tasks. More precisely,  \texttt{DrBERT-FS} achieves the highest performance in 8 tasks, \texttt{DrBERT-CP} in 5 tasks, and \texttt{CamemBERT-bio} in 2 tasks. This indicates that domain and language-specialized models achieve the best performance in up to 75\% of the DrBenchmark downstream tasks.

\begin{figure*}[t]
\centering
\subfloat[MorFITT CLS.]{\label{4figs-c} \includegraphics[width=0.20\textwidth]{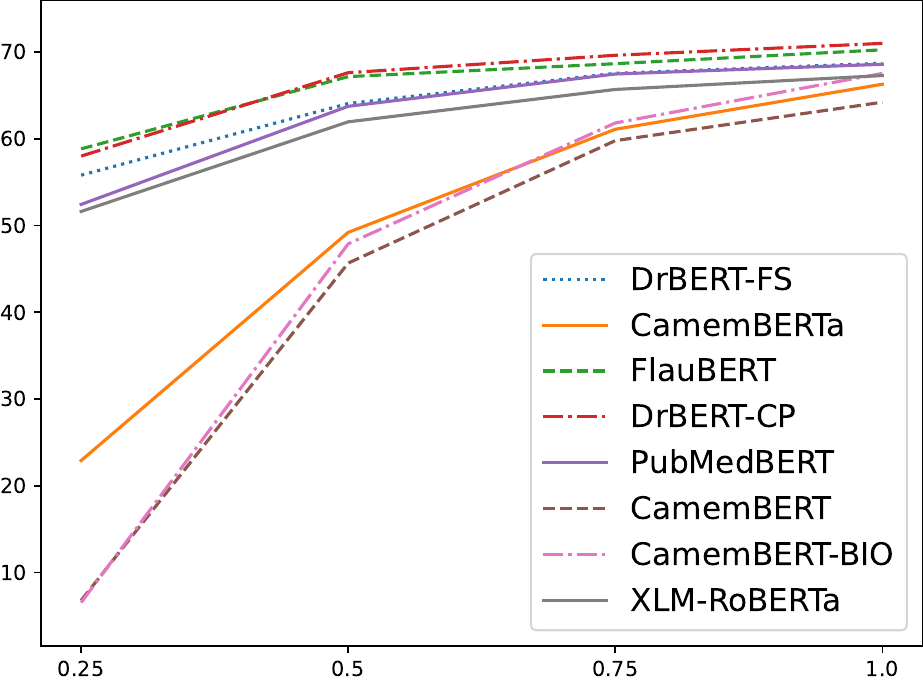}}%
\hfill
\subfloat[DEFT 2020 CLS]{\label{4figs-a} \includegraphics[width=0.20\textwidth]{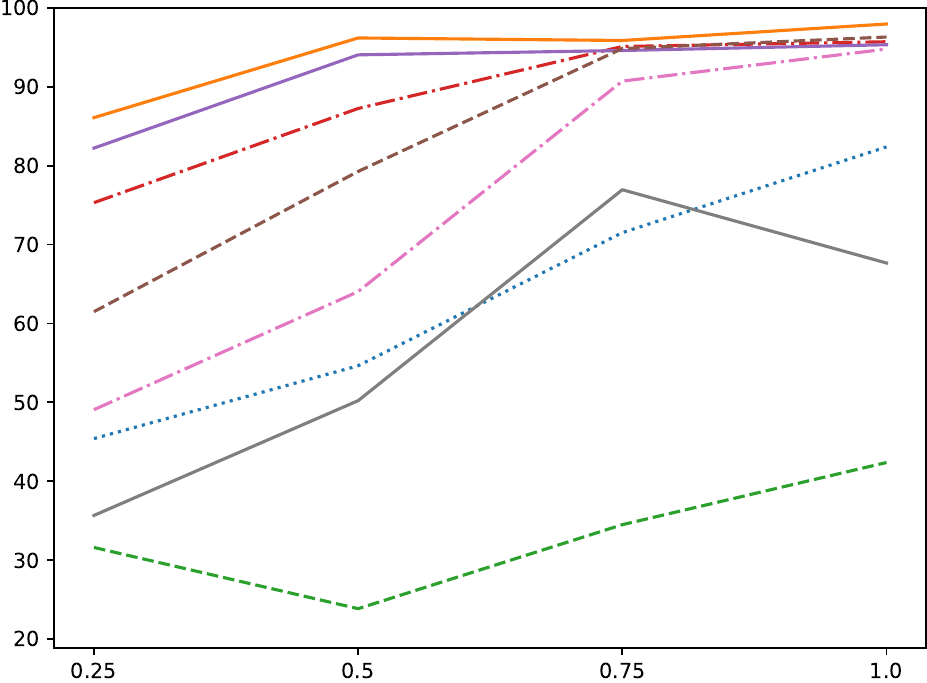}}
\hfill
\subfloat[E3C NER Clinical.]{\label{4figs-b} \includegraphics[width=0.20\textwidth]{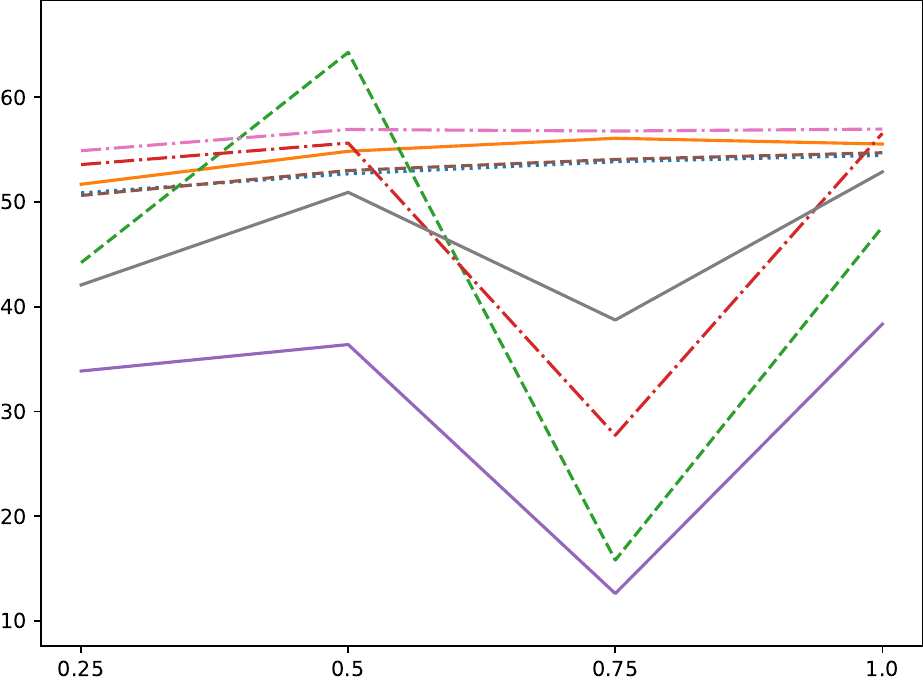}}%
\hfill
\subfloat[MantraGSC NER Medline.]{\label{4figs-d} \includegraphics[width=0.20\textwidth]{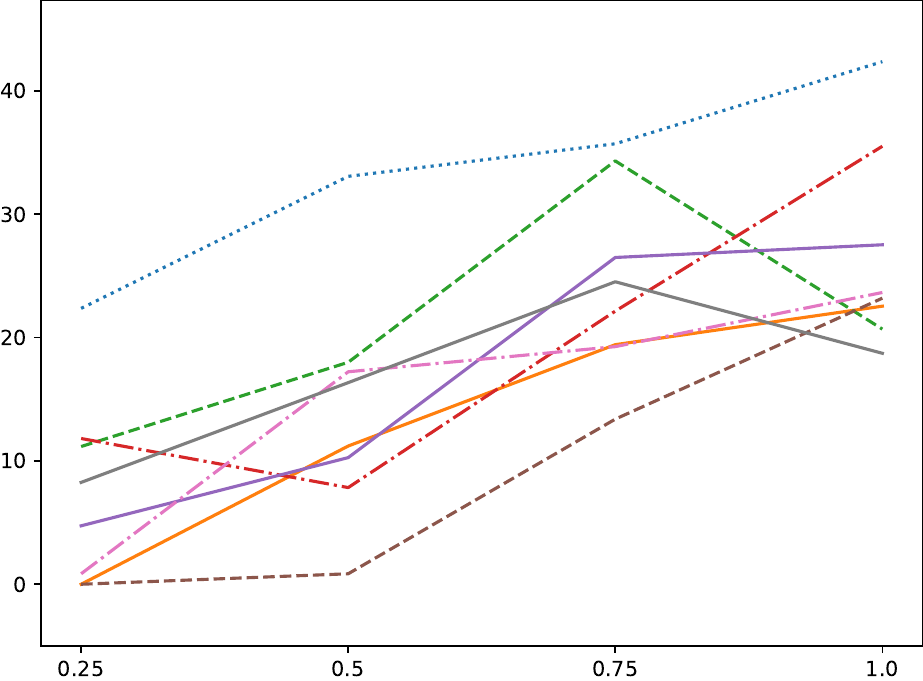}}%
\\
\subfloat[FrenchMedMCQA CLS]{\label{4figs-e} \includegraphics[width=0.20\textwidth]{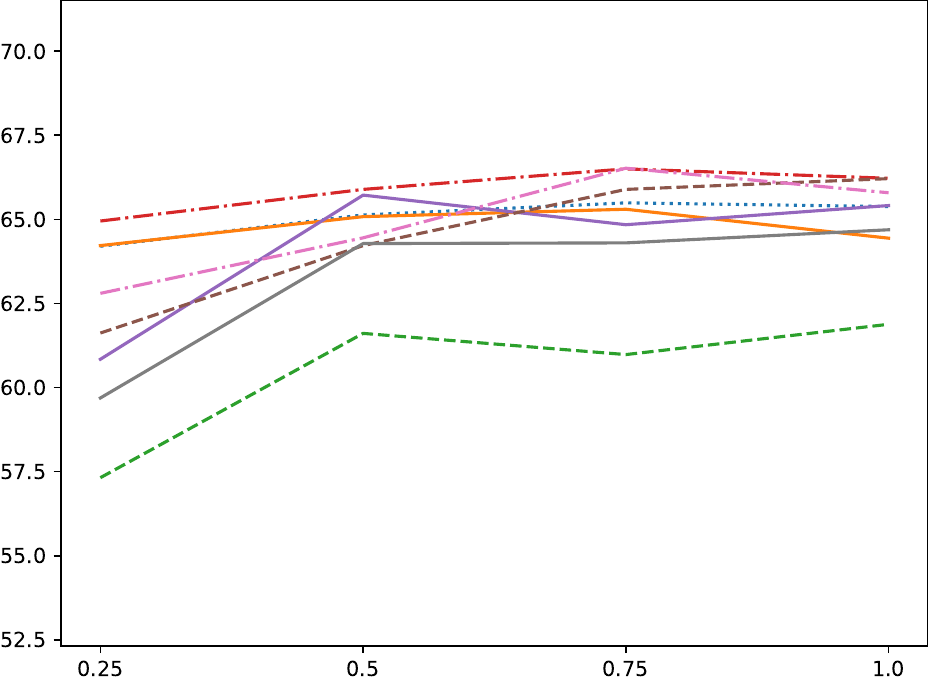}}
\hfill
\subfloat[MantraGSC NER EMEA.]{\label{4figs-f} \includegraphics[width=0.20\textwidth]{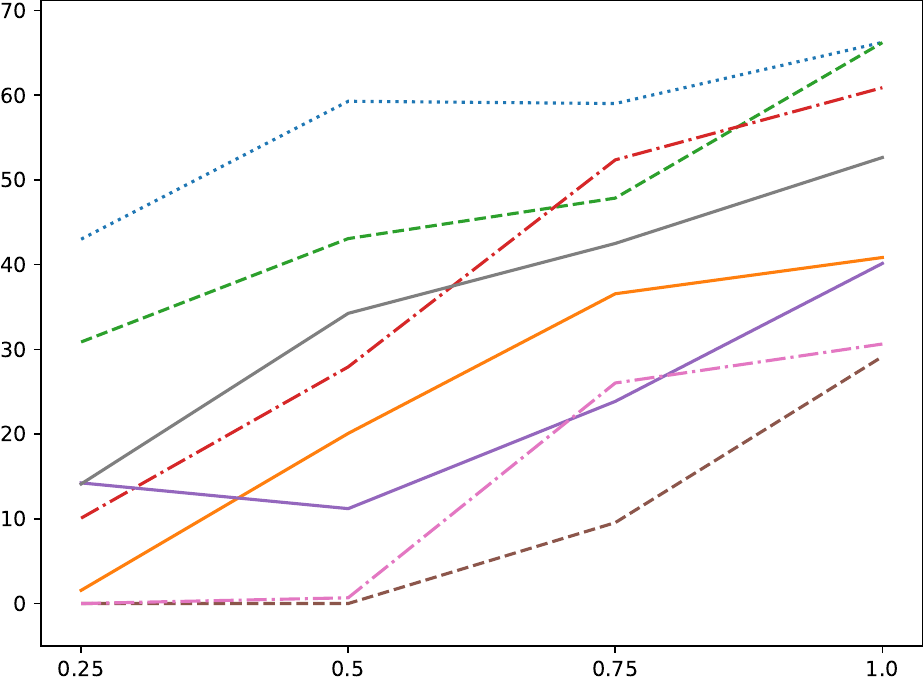}}%
\hfill
\subfloat[QUAERO NER EMEA.]{\label{4figs-g} \includegraphics[width=0.20\textwidth]{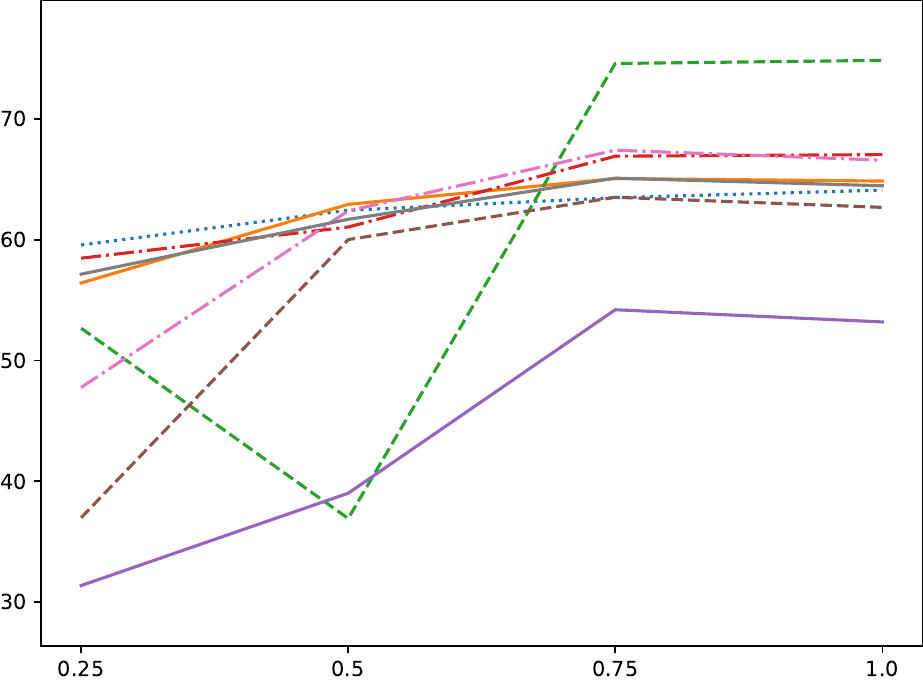}}%
\hfill
\subfloat[CLISTER STS.]{\label{4figs-h} \includegraphics[width=0.20\textwidth]{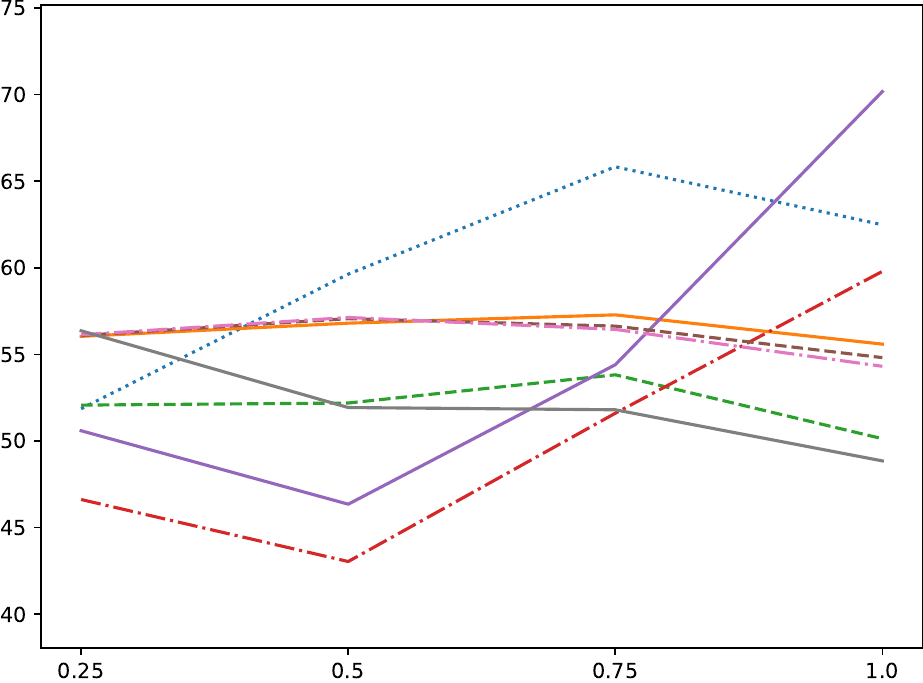}}%
\caption{Performance with varying training subset sizes (25\%, 50\%, 75\% and 100\%). Results are reported on the full test set.}
\label{4figs}
\end{figure*}

\paragraph{Biomedical vs. Generalist.} The nature of the data appears to have an influence. Generalist models (\texttt{CamemBERT}, \texttt{CamemBERTa}, \texttt{FlauBERT} and \texttt{XLM-RoBERTa}) are more suitable for tasks that require extensive linguistic knowledge but may not perform as well as specialized models nor even reach their level of performance. We observe that all generalist models obtain better performance only on 4 out of the 20 tasks, but still remain competitive on most tasks. Furthermore, our experiments with \texttt{DrBERT-FS} indicate that biomedical models may require less pre-training data compared to generalist ones. However, it is important to note that this observation requires further confirmation. In some tasks, biomedical models that undergo continual pre-training from a generalist model, such as \texttt{CamemBERT-bio}, can prove to be the most effective, underscoring the value of pre-training on generalist datasets.

\paragraph{From-scratch vs. Continual Pre-Training.} \texttt{DrBERT-CP} and \texttt{CamemBERT-bio}, pre-trained from \texttt{PubMedBERT} and \texttt{CamemBERT} respectively, demonstrate improved performance compared to their initial models. Notably, \texttt{DrBERT-CP} outperforms \texttt{CamemBERT-bio} in 15 out of 20 tasks. These findings suggest that when it comes to continual pre-training, starting with a specialized model in the specific domain (here, \texttt{PubMedBERT}) may be a better choice than a generalist model (here, \texttt{CamemBERT}), even with different languages. 
Additionally, we observe that \texttt{DrBERT-FS} achieves the highest performance in 8 tasks, suggesting that starting {\it from-scratch} can be a competitive strategy compared to {\it continual pre-training}.

\paragraph{French vs. Other language.} French models generally achieve better performance compared to English or multilingual ones. When considering the English \texttt{PubMedBERT} model, we observe that its performance in most tasks is comparable to that of the French models, with the exception of NER tasks where French models demonstrate superiority. Thus, we observe that the language appears to be less prominent when utilized in domain-specific tasks, such as those in the biomedical field.

\paragraph{RoBERTa vs. DeBERTaV3 architectures.} Despite being trained on only 30\% of the pre-training data used by \texttt{CamemBERT$_{CCNET}$}, \texttt{CamemBERTa} achieves identical or better performances in 68\% of the tasks (12 out of 20), benefiting from the DeBERTaV3 architecture in domain-specific scenarios. However, all the models based on \texttt{CamemBERT} face difficulties in corpora with limited amount of data, such as MantraGSC Patents, where they fail to generate labels other than 'O'. On the other hand, in the same low-resource scenarios, \texttt{CamemBERTa} models exhibit greater robustness and achieve superior performance. The architecture on which the models are based therefore seems to play a role in the performance obtained.

\begin{table*}[!htb]
\setlength\tabcolsep{3pt}
\resizebox{\textwidth}{!}{%
\setlength\extrarowheight{2.5pt}
\tiny
\centering
\begin{tabular}{ll|ccc|ccc|c|c}
\hline
&  & \multicolumn{3}{c|}{\textbf{French Generalist}} &
\multicolumn{3}{c|}{\textbf{French Biomedical}} & 
\textbf{English Biomedical} & 
\textbf{Cross-lingual Generalist} \\ \hline

\textbf{Dataset} & 
\textbf{Task} & 
\textbf{CamemBERT} & 
\textbf{CamemBERTa} & 
\textbf{FlauBERT} & 
\textbf{DrBERT-FS} & 
\textbf{DrBERT-CP} & 
\textbf{CamemBERT-bio} & 
\textbf{PubMedBERT} & 
\textbf{XLM-RoBERTa} \\ \hline

CAS
& POS & 1.63 & 1.64 & \textbf{1.34} & \cellcolor{green!25}1.36 & \underline{1.81} & 1.63 & \cellcolor{red!25}\underline{1.81} & 1.8 \\

\hline

ESSAI
& POS & 1.55 & 1.56 & \cellcolor{red!25}\textbf{1.28} & \cellcolor{green!25}1.29 & \underline{1.78} & 1.55 & \underline{1.78} & 1.75 \\
\hline

\multirow{2}{*}{QUAERO} 
& NER EMEA & 1.66 & 1.67 & \cellcolor{green!25} \textbf{1.37} & \textbf{1.37} & 1.73 & 1.66 & \cellcolor{red!25}1.73 & \underline{1.77} \\
& NER Medline & 2.01 & 2.01 & \cellcolor{red!25}\textbf{1.58} & 1.64 & \cellcolor{green!25} 1.97 & 2.01 & 1.97 & \underline{2.18} \\ \hline

\multirow{2}{*}{E3C}
& NER FR Clinical & 1.64 & 1.65 & 1.39 & \textbf{1.32} & \underline{1.80} & \cellcolor{green!25} 1.64 & \cellcolor{red!25}\underline{1.80} & 1.78 \\
& NER FR Temporal & \cellcolor{green!25} 1.63 & 1.63 & \cellcolor{red!25}1.38 & \textbf{1.31} & \underline{1.80} & 1.63 & \underline{1.80} & 1.76 \\ \hline

MorFITT & Multi-Label CLS & \cellcolor{red!25}1.51 & 1.51 & \textbf{1.33} & 1.39 & \cellcolor{green!25} \underline{1.91} & 1.51 & \underline{1.91} & 1.73 \\ \hline

\multirow{2}{*}{FrenchMedMCQA} 
& MCQA & 1.80 & 1.80 & \cellcolor{red!25}\textbf{1.55} & \textbf{1.55} & \underline{2.03} & \cellcolor{green!25} 1.80 & \underline{2.03} & 2.00 \\
& CLS & 1.80 & 1.80 & \cellcolor{red!25}\textbf{1.55} & \textbf{1.55} & \cellcolor{green!25} \underline{2.03} & 1.80 & \underline{2.03} & 2.00 \\ \hline

\multirow{3}{*}{MantraGSC} 
& NER FR EMEA & \cellcolor{red!25}1.50 & 1.46 & \textbf{1.34} & \cellcolor{green!25} 1.37 & \underline{1.99} & 1.50 & \underline{1.99} & 1.71 \\
& NER FR Medline & 2.25 & 2.25 & \textbf{1.88} & \cellcolor{green!25} 2.05 & 2.47 & 2.25 & 2.47 & \cellcolor{red!25}\underline{2.49} \\
& NER FR Patents & \cellcolor{red!25}1.58 & 1.58 & \textbf{1.41} & \cellcolor{green!25} 1.51 & \underline{2.06} & \cellcolor{red!25}1.58 & \underline{2.06} & 1.86 \\ \hline

CLISTER & STS & 1.76 & 1.76 & \textbf{1.55} & \textbf{1.55} & \underline{2.09} & 1.76 & \cellcolor{green!25} \underline{2.09} & \cellcolor{red!25}1.93 \\ \hline

\multirow{2}{*}{DEFT-2020} 
& STS & 1.43 & 1.43 & \cellcolor{red!25}\textbf{1.31} & 1.45 & \underline{1.92} & \cellcolor{red!25}1.43 & \cellcolor{green!25} \underline{1.92} & 1.64 \\
& CLS & 1.31 & \cellcolor{green!25} 1.32 & \cellcolor{red!25}\textbf{1.20} & 1.23 & \underline{1.75} & 1.31 & \underline{1.75} & 1.51 \\ \hline

\multirow{2}{*}{DEFT-2021} 
& CLS & 1.70 & 1.71 & \cellcolor{green!25} \textbf{1.48} & 1.51 & \underline{2.05} & \cellcolor{red!25}1.70 & \underline{2.05} & 1.90 \\
& NER & 1.62 & 1.63 & \cellcolor{red!25} \textbf{1.35} & \textbf{1.35} & \underline{1.80} & \cellcolor{green!25} 1.62 & \underline{1.80} & 1.79 \\ \hline

DiaMED & CLS & 1.66 & \cellcolor{red!25}1.67 & \textbf{1.45} & \cellcolor{green!25} 1.46 & \underline{1.99} & 1.66 & \underline{1.99} & 1.88 \\ \hline

\multirow{2}{*}{PxCorpus} 
& NER & 1.71 & 1.76 & \cellcolor{red!25}\textbf{1.63} & \cellcolor{green!25} 1.66 & \underline{2.13} & 1.71 & \underline{2.13} & 1.83 \\
& CLS & 1.71 & 1.76 & \textbf{1.63} & 1.66 & \cellcolor{green!25}\underline{2.13} & 1.71 & \cellcolor{red!25}\underline{2.13} & 1.83 \\ \hline\hline

\textit{Average} &  & \textit{1.67} & \textit{1.67} & \textbf{\textit{1.43}} & \textit{1.47} & \underline{\textit{1.90}} & \textit{1.67} & \underline{\textit{1.90}} & \textit{1.85} \\ \hline

\end{tabular}
}
\caption{Average sub-word units per word for each model and dataset. For each task, the lowest sub-word value is shown in bold, and the highest value is underlined. Models are grouped based on their tokenizer type. Cells in green indicate the best model in terms of performance for the task, while cells in red indicate the worst model.}
\label{table:avg-tokens-models}
\end{table*}

\subsection{Impact of fine-tuning with limited data}
\label{s:fine-tune}

Unlike the process of training language models, the fine-tuning approach involves utilizing annotated data to adapt a pre-trained language model for solving specific downstream tasks. In the previous section, we observed that language models pre-trained on medical data generally achieved better performance on DrBenchmark compared to generalist models trained on much larger datasets. However, we now question the models' ability to be effectively applied to biomedical tasks when there is limited fine-tuning training data available.
For this purpose, we conducted experiments by varying the amount of training data during the fine-tuning process by randomly choosing four percentages of the training data: 25\%, 50\%, 75\% and 100\%. To make the experiment as fair as possible, we did four runs for each percentage, model and dataset combination. The validation and test sets have not been changed for the sake of comparison. 

We observe that on certain datasets, some models capture information more quickly than others, like in Figures~\ref{4figs-a}, \ref{4figs-f} and \ref{4figs-c}.
Unsurprisingly, in almost all scenarios, having the complete training set yields better results than having only 25\% of it. However, we note few exceptions in Figures~\ref{4figs-c} and \ref{4figs-h} with \texttt{FlauBERT}, where we observe the opposite trend.
For intermediate percentages, 50\% and 75\%, we observe a decrease in performance with certain models, such as \texttt{FlauBERT} in Figures~\ref{4figs-c} and \ref{4figs-g}, and \texttt{DrBERT-CP} in Figures~\ref{4figs-d} and \ref{4figs-h}.
In NER tasks (Figures~\ref{4figs-c}, \ref{4figs-d}, \ref{4figs-f} and \ref{4figs-g}), \texttt{DrBERT-FS} achieves the best performance in scenarios with very little data, indicating good model robustness.

\subsection{Analysis of word tokenization}
\label{s:word_tok}

\begin{figure}[!htb]
\centering
\center
\includegraphics[scale=0.30]{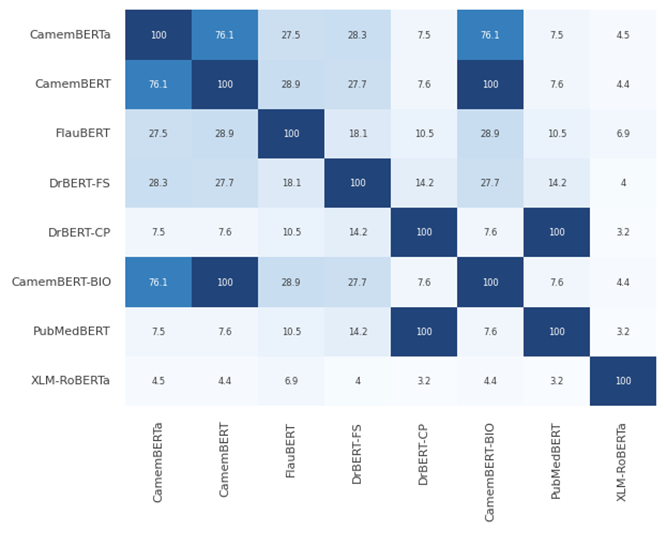}
\caption{Vocabularies inter-coverage matrix.}
\label{fig:Inter-Vocab-Coverage}
\end{figure}

Tokenizers play a crucial role in MLMs by utilizing size-limited vocabularies to split texts into sub-units, aiming to handle out-of-vocabulary (OOV) words. Due to variations in the training data, vocabularies differ across different models, as illustrated in Figure~\ref{fig:Inter-Vocab-Coverage}. As a result, tokenizers segment words in distinct ways, yet remarkably achieve similar performance levels as previously noted in Table~\ref{table:results}.

So far, there has been a prevailing notion in the community that excessive segmentation of words in tokenization leads to a loss of morphological form and semantic meaning, introducing noise and adversely affecting performance~\cite{church_2020,hofmann-etal-2021-superbizarre,bostrom-durrett-2020-byte}. However, our experiments, as shown in Table~\ref{table:avg-tokens-models}, reveal that \texttt{FlauBERT} is the model with the least word segmentation (1.43 in average), while \texttt{DrBERT-CP} tends to have the highest average segmentation (1.90 in average). Surprisingly, when comparing the performance of these two models on the benchmark tasks, we observe that \texttt{DrBERT-CP} outperforms \texttt{FlauBERT} on 16 out of the 20 tasks, thus contradicting previous conclusions drawn by the community. 
Table~\ref{table:sub-units} in Appendix \ref{sec:ann-tok} provides some examples of the tokenization done by each analyzed model, showcasing a list of commonly used biomedical terms. Yet, tokenization, as it is currently done in MLMs, seems to play a minor role in the performance of systems.

Table~\ref{t:avg_res} summarizes the results obtained on average by the considered MLMs when aggregating the tasks into one of the five designated categories: POS, NER, MCQA, MCC (Multi-class classification), MLC (Multi-label classification), or STS tasks. Upon analyzing the average performance by task category, it becomes evident that the leading model, \texttt{DrBERT-FS}, does not excel in tasks such as MLC or STS. For example, the multilingual biomedical model \texttt{PubMedBERT} demonstrates a notable advantage, with nearly 18 EDRM points ahead of \texttt{CamemBERT-bio} in the STS tasks.

\begin{table}[htb!]
\tiny
\centering
\setlength\extrarowheight{3pt}
\setlength\tabcolsep{5pt}
\begin{tabular}{ccccccc}
\hline
& \multicolumn{6}{c}{\textbf{Tasks}} \\ \hline
\multicolumn{1}{c|}{\textbf{Models}} & \textbf{POS} & \textbf{NER} & \textbf{MCQA} & \textbf{MCC*} & \textbf{MLC*} & \textbf{STS} \\ \hline
\multicolumn{1}{c|}{\textbf{CamemBERT}}     & 96.45 & 51.52 & 28.53 / 2.25 & 71.83 & 41.12 & 0.57 / 0.45 \\
\multicolumn{1}{c|}{\textbf{CamemBERTa}}    & 97.32 & 58.16 & 29.77 / 2.57 & 70.10 & 42.16 & 0.57 / 0.45 \\
\multicolumn{1}{c|}{\textbf{FlauBERT}}      & 96.13 & 51.85 & 27.88 / 2.09 & 57.94 & \textbf{54.73} & 0.54 / 0.40 \\
\multicolumn{1}{c|}{\textbf{DrBERT-FS}}     & \textbf{97.67} & \textbf{64.23} & 31.07 / \textbf{3.22} & 75.66 & \underline{51.42} & \underline{0.67} / \underline{0.69} \\
\multicolumn{1}{c|}{\textbf{DrBERT-CP}}     & 97.23 & \underline{59.84} & 32.41 / \underline{2.89} & \textbf{77.72} & 50.51 & 0.66 / 0.67 \\
\multicolumn{1}{c|}{\textbf{CamemBERT-bio}} & 96.30 & 53.06 & \textbf{35.30} / 1.45 & 73.65 & 42.67 & 0.56 / 0.29 \\
\multicolumn{1}{c|}{\textbf{PubMedBERT}}    & 96.12 & 46.93 & 32.90 / 1.61 & \underline{77.20} & 47.05 & \textbf{0.74} / \textbf{0.82} \\
\multicolumn{1}{c|}{\textbf{XLM-RoBERTa}}   & \underline{97.62} & 54.21 & \underline{34.74} / 2.09 & 63.23 & 45.87 & 0.54 / 0.24 \\ \hline
\end{tabular}%
\caption{Average results obtained by the different MLMs for each type of task. MLC stands for Multi-label classification and MCC for Multi-class classification.\label{t:avg_res}}
\end{table}

\section{Conclusion}
\label{sec:conclusion}

In this paper, we introduced DrBenchmark, the first large language understanding benchmark tailored for the French biomedical domain. We conducted a qualitative evaluation of 8 state-of-the-art masked language models (MLMs) on this comprehensive benchmark, encompassing 20 diverse downstream tasks. Our findings illuminate the limitations of generalist models in tackling complex biomedical tasks, emphasizing the importance of employing domain-specific models to achieve peak performance. While the French biomedical models excel in most tasks, no single model emerges as universally superior. Remarkably, certain out-of-domain models or models trained in different languages exhibit superior performance in specific tasks and maintain competitiveness in others.

In conclusion, we have observed that several biomedical tasks in DrBenchmark exhibit relatively poor performance, even when utilizing specialized biomedical models. We postulate that the models examined in this study, here state-of-the-art MLMs, may not be the most effective choices for specific tasks such as question-answering or multi-label classification. In our future research, we intend to shift our focus towards generative approaches, such as LLaMA~\cite{touvron2023llama}, OPT~\cite{zhang2022opt}, or GPT-NeoX-20B~\cite{black-etal-2022-gpt}, as well as their instruction-tuned counterparts~\cite{iyer2023optiml}. These alternatives may offer more suitable solutions for addressing these types of tasks.

\section{Acknowledgements}
\label{sec:Acknowledgements}

This work was performed using HPC resources from GENCI-IDRIS (Grant 2022-AD011013061R1 and 2022-AD011013715) and from CCIPL (Centre de Calcul Intensif des Pays de la Loire). This work was financially supported by ANR MALADES (ANR-23-IAS1-0005), ANR AIBy4 (ANR-20-THIA-0011) and Zenidoc.

\section*{Ethical considerations}
\label{sec:EthicalConsiderations}

All code for DrBenchmark is released under the MIT License. The licensing for all datasets remains unchanged from the original sources, and DrBenchmark has no intention of redistributing these datasets.

\section*{Limitations}
\label{sec:Limitations}


The quantitative study we conducted on the PLMs requires further in-depth analysis to comprehend the impact of different parameters. Firstly, we investigated the influence of tokenizers based on the average number of sub-tokens they produce per word. It is important to note that various tokenization algorithms are employed, depending on the model under examination. Therefore, it is necessary to specifically assess the impact of these algorithms on model construction. Additionally, the size of the data has not been thoroughly investigated, particularly the significance of the pre-training data size, especially specialized data in the biomedical field. Analyzing the influence and importance of the amount of data used would be crucial for gaining deeper insights.



Although the benchmark is easily reproducible and customizable, it required a substantial amount of computational power to execute all runs. We utilized approximately 2,500 hours on V100 GPUs from the Jean-Zay supercomputer to complete the quantitative study. According to the Jean Zay supercomputer documentation~\footnote{\href{http://www.idris.fr/media/jean-zay/jean-zay-conso-heure-calcul.pdf}{http://www.idris.fr/media/jean-zay/jean-zay-conso-heure-calcul.pdf}}, the total environmental cost is estimated to be equivalent to 647,500 Wh or 36.9 kgCO2eq/kWh, based on the carbon intensity of the energy grid mentioned in the BLOOM environmental cost study conducted on Jean Zay~\cite{luccioni2022estimating}. While we acknowledge the significant cost of our study, we believe it will enable the research community to direct their future studies more efficiently by providing a comprehensive overview of the performance and behavior of existing models. This will help prevent redundant evaluations of the same models.





\nocite{}
\section{Bibliographical References}\label{sec:reference}

\bibliographystyle{lrec-coling2024-natbib}
\bibliography{lrec-coling2024-example}


\section{Language Resource References}
\label{lr:ref}
\bibliographystylelanguageresource{lrec-coling2024-natbib}
\bibliographylanguageresource{languageresource}

\appendix

\section*{Appendices}

\section{Hyperparameters}
\label{sec:hyperparam}

For the experiments, we utilize the following hyperparameters that yield optimal performance from the models. To mitigate overfitting, we locally save the best model based on its validation metric.

\begin{table}[htb!]
\small
\centering
\begin{tabular}{lc}
\hline
Hyper-parameter & Value  \\
\hline
Max sequence length & 512 \\
Epochs & 20 \\
Batch size & 16 \\
Learning Rate & 2e-5 \\
Weight Decay & 0.01 \\
\hline
\end{tabular}
\caption{Hyper-parameters for the question-answering experiments.}
\label{table:hyperparam:qa}
\end{table}

\begin{table}[htb!]
\small
\centering
\begin{tabular}{lc}
\hline
Hyper-parameter & Value  \\
\hline
Max sequence length & 512 \\
Epochs & 10 / 25 / 35 \\
Batch size & 16 \\
Learning Rate & 2e-5 \\
Weight Decay & 0.01 \\
\hline
\end{tabular}
\caption{Hyper-parameters for the classification experiments. The number of epochs is by default 10 except for DEFT-2020 (25 epochs) and MorFITT (35 epochs).}
\label{table:hyperparam:cls}
\end{table}

\begin{table}[htb!]
\small
\centering
\begin{tabular}{lc}
\hline
Hyper-parameter & Value  \\
\hline
Max sequence length & 512 \\
Epochs & 10 \\
Batch size & 16 \\
Learning Rate & 1e-5 \\
Weight Decay & 0.01 \\
\hline
\end{tabular}
\caption{Hyper-parameters for the POS tagging experiments.}
\label{table:hyperparam:pos}
\end{table}

\begin{table}[htb!]
\small
\centering
\begin{tabular}{lc}
\hline
Hyper-parameter & Value  \\
\hline
Max sequence length & 512 \\
Epochs & 30 \\
Batch size & 16 \\
Learning Rate & 2e-5 \\
Weight Decay & 0.01 \\
\hline
\end{tabular}
\caption{Hyper-parameters for the regression experiments.}
\label{table:hyperparam:regr}
\end{table}

\begin{table}[htb!]
\small
\centering
\begin{tabular}{lc}
\hline
Hyper-parameter & Value  \\
\hline
Max sequence length & 512 \\
Epochs & 15 \\
Batch size & 16 \\
Learning Rate & 1e-4 \\
Weight Decay & 0.01 \\
\hline
\end{tabular}
\caption{Hyper-parameters for the NER experiments.}
\label{table:hyperparam:ner}
\end{table}

\newpage

\onecolumn

\section{Dataset Classes}
\label{sec:classes}

\subsection{CAS}
\label{sec:classes-CAS}

\textit{INT}, \textit{PRO:DEM}, \textit{VER:impf}, \textit{VER:ppre}, \textit{PRP:det}, \textit{KON}, \textit{VER:pper}, \textit{PRP}, \textit{PRO:IND}, \textit{VER:simp}, \textit{VER:con}, \textit{SENT}, \textit{VER:futu}, \textit{PRO:PER}, \textit{VER:infi}, \textit{ADJ}, \textit{NAM}, \textit{NUM}, \textit{PUN:cit}, \textit{PRO:REL}, \textit{VER:subi}, \textit{ABR}, \textit{NOM}, \textit{VER:pres}, \textit{DET:ART}, \textit{VER:cond}, \textit{VER:subp}, \textit{DET:POS}, \textit{ADV}, \textit{SYM} and \textit{PUN}.

\subsection{ESSAI}
\label{sec:classes-ESSAI}

\textit{INT}, \textit{PRO:POS}, \textit{PRP}, \textit{SENT}, \textit{PRO}, \textit{ABR}, \textit{VER:pres}, \textit{KON}, \textit{SYM}, \textit{DET:POS}, \textit{VER:}, \textit{PRO:IND}, \textit{NAM}, \textit{ADV}, \textit{PRO:DEM}, \textit{NN}, \textit{PRO:PER}, \textit{VER:pper}, \textit{VER:ppre}, \textit{PUN}, \textit{VER:simp}, \textit{PREF}, \textit{NUM}, \textit{VER:futu}, \textit{NOM}, \textit{VER:impf}, \textit{VER:subp}, \textit{VER:infi}, \textit{DET:ART}, \textit{PUN:cit}, \textit{ADJ}, \textit{PRP:det}, \textit{PRO:REL}, \textit{VER:cond} and \textit{VER:subi}.

\subsection{QUAERO}
\label{sec:classes-QUAERO}

\textit{O}, \textit{GEOG}, \textit{PHEN}, \textit{DISO}, \textit{ANAT}, \textit{OBJC}, \textit{PHYS}, \textit{PROC}, \textit{DEVI}, \textit{CHEM} and \textit{LIVB}

\subsection{E3C}
\label{sec:classes-E3C}

\paragraph{Clinical:} 
\textit{O}, and \textit{CLINENTITY}
            
\paragraph{Temporal:} \textit{O}, \textit{EVENT}, \textit{ACTOR}, \textit{BODYPART}, \textit{TIMEX3} and \textit{RML}

\subsection{MorFITT}
\label{sec:classes-MorFITT}

\textit{microbiology}, \textit{etiology}, \textit{virology}, \textit{physiology}, \textit{immunology}, \textit{parasitology}, \textit{genetics}, \textit{chemistry}, \textit{veterinary}, \textit{surgery}, \textit{pharmacology} and \textit{psychology}

\subsection{MantraGSC}
\label{sec:classes-MantraGSC}

\paragraph{Medline:} \textit{ANAT}, \textit{PROC}, \textit{CHEM}, \textit{PHYS}, \textit{GEOG}, \textit{DEVI}, \textit{LIVB}, \textit{OBJC}, \textit{DISO}, \textit{PHEN} and \textit{O}.

\paragraph{EMEA and Patents:} \textit{ANAT}, \textit{PROC}, \textit{CHEM}, \textit{PHYS}, \textit{DEVI}, \textit{LIVB}, \textit{OBJC}, \textit{DISO}, \textit{PHEN} and \textit{O}.

\subsection{DEFT-2021}
\label{sec:classes-DEFT2021}

\paragraph{Multi-label Classification:} \textit{immunitaire (immunology)}, \textit{endocriniennes (endocrinology)}, \textit{blessures (injury)}, \textit{chimiques (chemicals)}, \textit{etatsosy (signs and symptoms)}, \textit{nutritionnelles (nutrition)}, \textit{infections (infections)}, \textit{virales (virology)}, \textit{parasitaires (parasitology)}, \textit{tumeur (oncology)}, \textit{osteomusculaires (osteomuscular disorders)}, \textit{stomatognathique (stomatology)}, \textit{digestif (digestive system disorders)}, \textit{respiratoire (respiratory system disorders)}, \textit{ORL (otorhinolaryngologic diseases)}, \textit{nerveux (nervous system disorders)}, \textit{oeil (eye diseases)}, \textit{homme (male genital diseases)}, \textit{femme (female genital diseases)}, \textit{cardiovasculaires (cardiology)}, \textit{hemopathies (hemic and lymphatic diseases)}, \textit{genetique (genertic disorders)} and \textit{peau (dermatology)}.

\paragraph{Named-entity recognition:} \textit{O}, \textit{ANATOMY}, \textit{DATE}, \textit{DOSAGE}, \textit{DURATION}, \textit{MEDICAL EXAM}, \textit{FREQUENCY}, \textit{MODE}, \textit{MOMENT}, \textit{PATHOLOGY}, \textit{SOSY}, \textit{SUBSTANCE}, \textit{TREATMENT} and \textit{VALUE}

\subsection{DiaMed}
\label{sec:classes-DiaMed}

\begin{itemize}
\item \textit{A00-B99  Certain infectious and parasitic diseases}

\item \textit{C00-D49  Neoplasms}

\item \textit{D50-D89  Diseases of the blood and blood-forming organs and certain disorders involving the immune mechanism}

\item \textit{E00-E89  Endocrine, nutritional and metabolic diseases}

\item \textit{F01-F99  Mental, Behavioral and Neurodevelopmental disorders}

\item \textit{G00-G99  Diseases of the nervous system}

\item \textit{H00-H59  Diseases of the eye and adnexa}

\item \textit{H60-H95  Diseases of the ear and mastoid process}

\item \textit{I00-I99  Diseases of the circulatory system}

\item \textit{J00-J99  Diseases of the respiratory system}

\item \textit{K00-K95  Diseases of the digestive system}

\item \textit{L00-L99  Diseases of the skin and subcutaneous tissue}

\item \textit{M00-M99  Diseases of the musculoskeletal system and connective tissue}

\item \textit{N00-N99  Diseases of the genitourinary system}

\item \textit{O00-O9A  Pregnancy, childbirth and the puerperium}

\item \textit{P00-P96  Certain conditions originating in the perinatal period}

\item \textit{Q00-Q99  Congenital malformations, deformations and chromosomal abnormalities}

\item \textit{R00-R99  Symptoms, signs and abnormal clinical and laboratory findings, not elsewhere classified}

\item \textit{S00-T88  Injury, poisoning and certain other consequences of external causes}

\item \textit{U00-U85  Codes for special purposes}

\item \textit{V00-Y99  External causes of morbidity}

\item \textit{Z00-Z99  Factors influencing health status and contact with health services}

\end{itemize}

\subsection{PxCorpus}
\label{sec:classes-pxcorpus}

\paragraph{Intent classification:} \textit{MEDICAL PRESCRIPTION}, \textit{NEGATE}, \textit{NONE} and \textit{REPLACE}

\paragraph{Named-entity recognition:} \textit{O}, \textit{A}, \textit{CMA\_EVENT}, \textit{D\_DOS\_FORM}, \textit{D\_DOS\_FORM\_EXT}, \textit{D\_DOS\_UP}, \textit{D\_DOS\_VAL}, \textit{DOS\_COND}, \textit{DOS\_UF}, \textit{DOS\_VAL}, \textit{DRUG}, \textit{DUR\_UT}, \textit{DUR\_VAL}, \textit{FASTING}, \textit{FREQ\_DAYS}, \textit{FREQ\_INT\_V1}, \textit{FREQ\_INT\_V1\_UT}, \textit{FREQ\_INT\_V2}, \textit{FREQ\_INT\_V2\_UT}, \textit{FREQ\_STARTDAY}, \textit{FREQ\_UT}, \textit{FREQ\_VAL}, \textit{INN}, \textit{MAX\_UNIT\_UF}, \textit{MAX\_UNIT\_UT}, \textit{MAX\_UNIT\_VAL}, \textit{MIN\_GAP\_UT}, \textit{MIN\_GAP\_VAL}, \textit{QSP\_UT}, \textit{QSP\_VAL}, \textit{RE\_UT}, \textit{RE\_VAL}, \textit{RHYTHM\_HOUR}, \textit{RHYTHM\_PERDAY}, \textit{RHYTHM\_REC\_UT}, \textit{RHYTHM\_REC\_VAL}, \textit{RHYTHM\_TDTE} and \textit{ROA}

\section{Word tokenization}
\label{sec:ann-tok}

\begin{table*}[htb]
\centering
\setlength\extrarowheight{4pt}
\resizebox{\textwidth}{!}{%
\begin{tabular}{c|ccc|c|c|c}
\hline

&
\multicolumn{3}{c|}{\textbf{French Generalist}} &
\textbf{French Biomedical} &
\textbf{English Biomedical} &
\textbf{Cross-lingual Generalist} \\ \hline

\multirow{2}{*}{\textbf{Term}} &
\multirow{2}{*}{\textbf{CamemBERTa}} &
\textbf{CamemBERT} &
\multirow{2}{*}{\textbf{FlauBERT}} &
\multirow{2}{*}{\textbf{DrBERT-FS}} &
\textbf{PubMedBERT} &
\multirow{2}{*}{\textbf{XLM-RoBERTa}} \\

& & \textbf{CamemBERT-bio} & & & \textbf{DrBERT-CP} & \\ \hline

\textit{asymptomatique} & a-s-ym-pto-matique & a-s-y-mp-to-matique & as-ym-ptom-atique & \checkmark & asympt-omat-ique & as-y-mp-tomat-ique \\
\textit{blépharorraphie} & blé-phar-or-ra-phi-e & blé-phar-or-ra-phi-e & bl-é-phar-or-raph-ie & blé-ph-ar-or-ra-ph-ie & ble-pha-ror-ra-phi-e & b-lép-har-orra-phi-e \\
\textit{bradycardie} & brad-y-cardi-e & brad-y-cardi-e & bra-dy-car-die & \checkmark & brady-car-di-e & bra-dy-card-ie \\
\textit{bronchographie} & bronch-ographie & bron-ch-ographie & bron-cho-graphie & bronch-ographie & bronch-ograph-ie & bron-ch-ographie \\
\textit{bronchopneumopathie} & bronch-op-ne-um-opathie & bron-cho-p-ne-um-opathie & bron-chop-neu-mo-pathie & bronchop-neumopathie & bronch-op-neum-opath-ie & bron-chop-ne-umo-pathi-e \\
\textit{dysménorrhée} & dys-mén-or-r-h-ée & dys-mén-or-r-h-ée & dys-mé-nor-rh-ée & dys-m-énorrhée & dysm-eno-rr-he-e & dys-mén-or-r-hé-e \\
\textit{glaucome} & gla-uc-ome & gla-uc-ome & glau-come & \checkmark & glauc-ome & gla-u-come \\
\textit{IRM} & \checkmark & \checkmark & \checkmark & \checkmark & ir-m & I-RM \\
\textit{kystectomie} & k-yst-ectomie & ky-st-ectomie & ky-st-ec-tomie & kys-tectomie & ky-st-ectom-ie & ky-st-ecto-mie \\
\textit{neuroleptique} & neuro-le-p-tique & neuro-le-p-tique & neur-ol-ep-tique & neur-oleptique & neurol-ept-ique & neuro-lep-tique \\
\textit{nicotine} & \checkmark & \checkmark & \checkmark & \checkmark & \checkmark & nico-tine \\
\textit{poliomyélite} & poli-om-y-élite & poli-om-y-élite & poli-omy-élite & poli-omyélite & poli-omyel-ite & poli-om-y-é-lite \\
\textit{rhinopharyngite} & rh-ino-phar-y-ng-ite & rhin-oph-ary-ng-ite & rh-ino-phar-yn-gite & rhin-opharyng-ite & rhin-oph-aryng-ite & r-hin-op-har-y-ng-ite \\
\textit{toxicomanie} & toxico-mani-e & toxico-mani-e & \checkmark & \checkmark & toxic-oman-ie & toxic-om-anie \\
\textit{vasoconstricteur} & vas-oc-on-strict-eur & vas-oc-on-strict-eur & vas-o-cons-tri-cteur & vasoconstric-teur & vasoconstric-te-ur & vaso-con-strict-eur \\

\hline
\end{tabular}%
}
\caption{Visual comparison of models' tokenization on commonly used biomedical terms. A checkmark indicates that the word is present as a complete token, while hyphens separate subword units.}
\label{table:sub-units}
\end{table*}

\end{document}